\documentclass[10pt,journal,compsoc]{IEEEtran}
\ifCLASSOPTIONcompsoc
  \usepackage[nocompress]{cite}
\else
  \usepackage{cite}
\fi
\ifCLASSINFOpdf
  \usepackage[pdftex]{graphicx}
\else
\fi
\usepackage{amsmath}
\usepackage{amssymb}
\usepackage{url}

\usepackage{multirow}
\usepackage{booktabs}
\usepackage{amsfonts}
\usepackage{amsthm}

\usepackage{hyperref}

\usepackage{color}
\definecolor{light}{rgb}{0.3, 0.3, 0.3}
\def\light#1{{\color{light}#1}}

\definecolor{my_color}{rgb}{0, 0, 0}

\usepackage{xcolor,colortbl}

\newcolumntype{a}{>{\columncolor{red}}c}

\begin{document}
%
\title{Non-Local Graph Neural Networks}
%
%
%
%

\author{Meng Liu*, Zhengyang~Wang*,
    and~Shuiwang~Ji,~\IEEEmembership{Senior~Member,~IEEE}
    \IEEEcompsocitemizethanks{\IEEEcompsocthanksitem Meng Liu and Shuiwang Ji are with the Department of Computer Science and Engineering, Texas A\&M University, College Station, TX 77843, USA. (E-mail: sji@tamu.edu)\protect
    \IEEEcompsocthanksitem Zhengyang Wang was with the Department of Computer Science and Engineering,
Texas A\&M University, and is currently with Amazon.com Services LLC.
    \IEEEcompsocthanksitem Meng Liu and Zhengyang Wang contribute equally to this work, and their names are in alphabetical order.
}
}

\IEEEtitleabstractindextext{%
\begin{abstract}
Modern graph neural networks (GNNs) learn node embeddings through multilayer local aggregation and achieve great success in applications on assortative graphs. However, tasks on disassortative graphs usually require non-local aggregation. In addition, we find that local aggregation is even harmful for some disassortative graphs. In this work, we propose a simple yet effective non-local aggregation framework with an efficient attention-guided sorting for GNNs. Based on it, we develop various non-local GNNs. We perform thorough experiments to analyze disassortative graph datasets and evaluate our non-local GNNs. Experimental results demonstrate that our non-local GNNs significantly outperform previous state-of-the-art methods on \textcolor{my_color}{seven} benchmark datasets of disassortative graphs, in terms of both model performance and efficiency.
\end{abstract}

}

\maketitle

\IEEEdisplaynontitleabstractindextext

%
\IEEEpeerreviewmaketitle

\IEEEraisesectionheading{\section{Introduction}\label{sec:intro}}

%
%
%
%
\IEEEPARstart{G}{raph} neural networks~(GNNs) process graphs and map each node to an embedding vector~\cite{bronstein2017geometric,gilmer2017neural,battaglia2018relational,zhou2018graph,zhang2018graph,wu2019comprehensive,zhang2020deep}. These node embeddings can be directly used for node-level applications, such as node classification~\cite{kipf2016semi}, link prediction~\cite{zhang2018link}. In addition, they can be used to learn the graph representation vector with graph pooling~\cite{ying2018hierarchical,zhang2018end,gao2019graph,ma2019graph,lee2019self,Yuan2020StructPool:,bianchi2020spectral}, in order to fit graph-level tasks~\cite{yanardag2015deep}. Many variants of GNNs have been proposed, such as ChebNets~\cite{defferrard2016convolutional}, GCNs~\cite{kipf2016semi}, GraphSAGE~\cite{hamilton2017inductive}, GATs~\cite{velivckovic2017graph}, LGCN~\cite{gao2018large} and GINs~\cite{xu2018powerful}. Their advantages have been shown on various graph datasets and tasks~\cite{errica2019fair}. However, these GNNs share a multilayer local aggregation framework, which is similar to convolutional neural networks~(CNNs)~\cite{lecun1998gradient} on grid-like data such as images and texts.

The importance of non-local aggregation has been recently demonstrated in many applications in the field of computer vision~\cite{wang2018non,wang2020non} and natural language processing~\cite{vaswani2017attention}. In particular, the attention mechanism has been widely explored to achieve non-local aggregation and capture long-range dependencies from distant locations. Basically, the attention mechanism measures the similarity between every pair of locations and enables information to be communicated among distant but similar locations. In terms of graphs, non-local aggregation is also crucial for disassortative graphs in which nodes with the same label are distant from each other, while previous studies of GNNs focus on assortative graph datasets (Section~\ref{sec:why_non_local}). In addition, we find that local aggregation is even harmful for some disassortative graphs (Section~\ref{sec:exp_dis}). The recently proposed Geom-GCN~\cite{pei2020geom} explores to capture long-range dependencies in disassortative graphs. It contains an attention-like step that computes the Euclidean distance between every pair of nodes. However, this step is computationally prohibitive for large-scale graphs, as the computational complexity is quadratic in the number of nodes. In addition, Geom-GCN employs pre-trained node embeddings~\cite{tenenbaum2000global,nickel2017poincare,ribeiro2017struc2vec} that are not task-specific, limiting the effectiveness and flexibility.

In this work, we propose a simple yet effective non-local aggregation framework for GNNs. At the heart of the framework lies an efficient attention-guided sorting, which enables non-local aggregation through classic local aggregation operators in general deep learning. The proposed framework can be flexibly used to augment common GNNs with low computational costs. Based on the framework, we build various efficient non-local GNNs. In addition, we perform detailed analysis on existing disassortative graph datasets, and apply different non-local GNNs accordingly. Experimental results show that our non-local GNNs significantly outperform previous methods on node classification tasks on \textcolor{my_color}{seven} benchmark datasets of disassortative graphs.

\section{Background and Related Work}\label{sec:related}

\subsection{Graph Neural Networks}\label{sec:gnn}

We focus on learning the embedding vector for each node through graph neural networks~(GNNs). Most existing GNNs follow a local aggregation framework. In general, each layer of GNNs scans every node in the graph and aggregates local information from directly connected nodes, \emph{i.e.}, the 1-hop neighbors. Specifically, a common layer of GNNs performs a two-step processing similar to the depthwise separable convolution~\cite{chollet2017xception}: spatial aggregation and feature transformation. The first step updates each node embedding using embedding vectors of spatially neighboring nodes. For example, GCNs~\cite{kipf2016semi} and GATs~\cite{velivckovic2017graph} compute a weighted sum of node embeddings within the 1-hop neighborhood, where weights come from the degree of nodes and the interaction between nodes, respectively. GraphSAGE~\cite{hamilton2017inductive} applies the max pooling, while GINs~\cite{xu2018powerful} simply sums the node embeddings. The feature transformation step is similar to the $1 \times 1$ convolution, where each node embedding vector is mapped into a new feature space through a shared linear transformation~\cite{kipf2016semi,hamilton2017inductive,velivckovic2017graph} or multilayer perceptron~(MLP)~\cite{xu2018powerful}. Different from these studies, LGCN~\cite{gao2018large} explores to directly apply the regular convolution through top-$k$ ranking.

Nevertheless, each layer of these GNNs only aggregates local information within the 1-hop neighborhood. While stacking multiple layers can theoretically enable communication between nodes across the multi-hop neighborhood, \textcolor{my_color}{the receptive field of a multiple-layer GCN usually includes more noise than useful information.} In addition, deep GNNs usually suffer from the over-smoothing problem~\cite{xu2018representation,li2018deeper,chen2019measuring} and the over-squashing issue~\cite{alon2020bottleneck}.

\subsection{Assortative and Disassortative Graphs}\label{sec:why_non_local}

There are many kinds of graphs in the literature, such as citation networks~\cite{sen2008collective}, community networks~\cite{chen2019measuring}, co-occurrence networks~\cite{tang2009social}, and webpage linking networks~\cite{rozemberczki2019multi}. We focus on graph datasets for the node classification tasks. In particular, we categorize graph datasets into assortative and disassortative ones~\cite{newman2002assortative,ribeiro2017struc2vec} according to the node homophily in terms of labels, \emph{i.e.}, how likely nodes with the same label are near each other in the graph.

Assortative graphs refer to those with a high node homophily. Common assortative graph datasets are citation networks and community networks. On the other hand, graphs in disassortative graph datasets contain more nodes that have the same label but are distant from each other. Example disassortative graph datasets are co-occurrence networks and webpage linking networks.

As shown in~\cite{ribeiro2017struc2vec}, distant nodes with the same label in disassortative graphs could be structurally similar. Thus, they are informative to each other. Most existing GNNs perform local aggregation only and achieve good performance on assortative graphs~\cite{kipf2016semi,hamilton2017inductive,velivckovic2017graph,gao2018large}. However, they may fail on disassortative graphs, where informative nodes in the same class tend to be out of the local multi-hop neighborhood and non-local aggregation is needed. Thus, in this work, we explore the non-local GNNs.

\subsection{Attention Mechanism}\label{sec:attn}

The attention mechanism~\cite{vaswani2017attention} has been widely used in GNNs~\cite{velivckovic2017graph,gao2019graphattn,knyazev2019understanding} as well as other deep learning models~\cite{yang2016hierarchical,wang2018non,wang2020non}. A typical attention mechanism takes three groups of vectors as inputs, namely the query vector $q$, key vectors $(k_1, k_2, \ldots, k_n)$, value vectors $(v_1, v_2, \ldots, v_n)$. Note that key and value vectors have a one-to-one correspondence and can be the same sometimes. The attention mechanism computes the output vector $o$ as
\begin{equation}
\begin{aligned}
o &= \sum_i a_i v_i, \\
a_i &= \textsc{Attend}(q, k_i) \in \mathbb{R}, \  i=1,2,\ldots, n,
\end{aligned}
\end{equation}
where the $\textsc{Attend}(\cdot)$ function could be any function that outputs a scalar attention score $a_i$ from the interaction between $q$ and $k_i$, such as dot product~\cite{gao2019graphattn} or even a neural network~\cite{velivckovic2017graph}. The definition of the three groups of input vectors depends on the models and applications.

\begin{figure*}[th]
	\centering
	\includegraphics[width=0.9\textwidth]{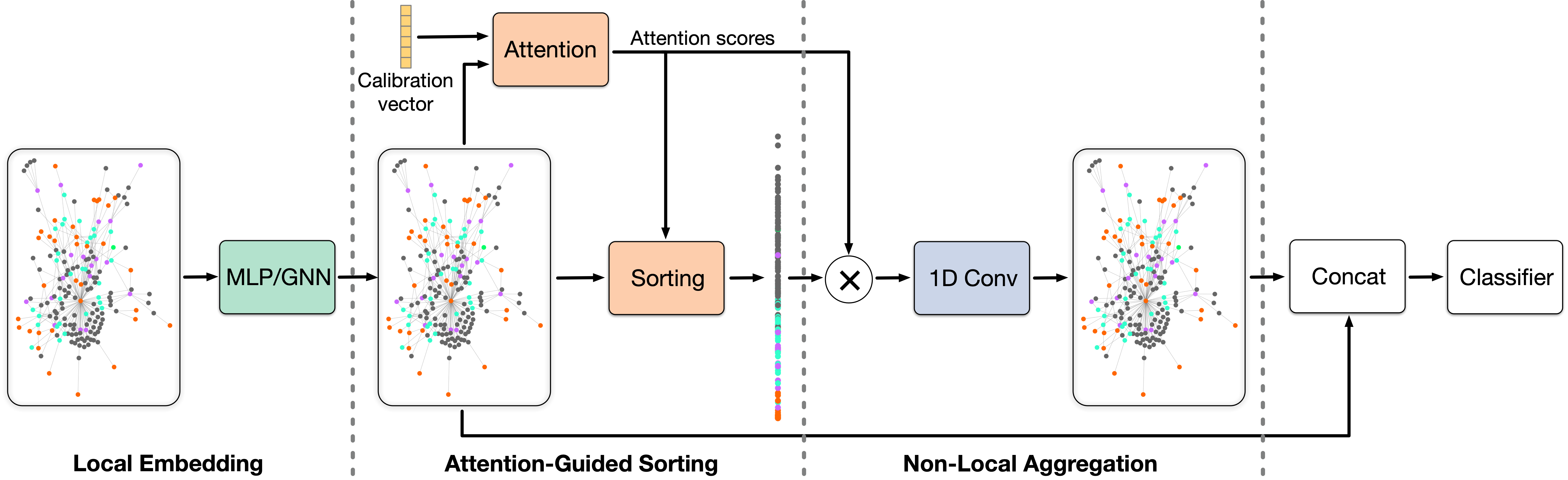}
	\vspace{-0.15in}
	\caption{An illustration of the proposed non-local aggregation framework. It consists of three steps, including local embedding, attention-guided sorting, and non-local aggregation. Details are described in Section~\ref{sec:framework}.}
	\label{fig:NLGNN}
\end{figure*}

Notably, existing GNNs usually use the attention mechanism for local aggregation~\cite{velivckovic2017graph,gao2019graphattn}. Specifically, when aggregating information for node $v$, the query vector is the embedding vector of $v$ while the key and value vectors come from node embeddings of $v$'s directly connected nodes. Note that the attention mechanism can be easily extended for non-local aggregation~\cite{wang2018non,wang2020non}, by letting the key and value vectors correspond to all the nodes in the graph when aggregating information for each node. However, it is computationally prohibitive given large-scale graphs, as iterating it for each node in a graph of $n$ nodes requires $O(n^2)$ time. In this work, we propose a novel non-local aggregation method that only requires $O(n\log n)$ time.

\section{The Proposed Method}\label{sec:method}

\subsection{Non-Local Aggregation with Attention-Guided Sorting}\label{sec:framework}

We consider a graph $\mathcal{G}=(V,E)$, where $V$ is the set of nodes and $E$ is the set of edges. Each edge $e \in E$ connects two nodes so that $E \subseteq V \times V$. Each node $v \in V$ has a node feature vector $x_v \in \mathbb{R}^d$. The $k$-hop neighborhood of $v$ refers to the set of nodes $\mathcal{N}_k(v)$ that can reach $v$ within $k$ edges. For example, the set of $v$'s directly connected nodes is its 1-hop neighborhood $\mathcal{N}_1(v)$.

As illustrated in Figure~\ref{fig:NLGNN}, our proposed non-local aggregation framework is composed of three steps, namely local embedding, attention-guided sorting, and non-local aggregation. In the following, we describe them one by one.

\textbf{Local Embedding:} Our proposed framework is built upon a local embedding step that extracts local node embeddings from the node feature vectors. The local embedding step can be as simple as
\begin{equation}
z_v = \textsc{MLP}(x_v) \in \mathbb{R}^{f},\ \forall v \in V.
\end{equation}
The $\textsc{MLP}(\cdot)$ function is a multilayer perceptron~(MLP), and $f$ is the dimension of the local node embedding $z_v$. Note that the $\textsc{MLP}(\cdot)$ function is shared across all the nodes in the graph. Applying MLP only takes the node itself into consideration without aggregating information from the neighborhood. This property is very important on some disassortative graphs, as shown in Section~\ref{sec:exp_dis}.

On the other hand, graph neural networks~(GNNs) can be used as the local embedding step as well, so that our proposed framework can be easily employed to augment existing GNNs. As introduced in Section~\ref{sec:gnn}, modern GNNs perform multilayer local aggregation. Typically, for each node, one layer of a GNN aggregates information from its 1-hop neighborhood. Stacking $L$ such local aggregation layers allows each node to access information that is $L$ hops away. To be specific, for each node $v \in V$, the $\ell$-th layer of a $L$-layer GNN $(\ell=1,2,\ldots,L)$ can be described as
\begin{equation}
z_v^{(\ell)} = \textsc{T}^{(\ell)} \left( \textsc{A}^{(\ell)} \left( \{ z_u^{(\ell-1)} : u \in \mathcal{N}_1(v) \cup v \} \right) \right) \in \mathbb{R}^{f},
\end{equation}
where $z_v^{(0)} = x_v$, and $z_v = z_v^{(L)}$ is the local node embedding. $\textsc{A}^{(\ell)}(\cdot)$ and $\textsc{T}^{(\ell)}(\cdot)$ functions represent the spatial aggregation and feature transformation step introduced in Section~\ref{sec:gnn}, respectively. With the above framework, GNNs can capture the node feature information from nodes within a local neighborhood as well as the structural information.

When either MLP or GNNs is used as the local embedding step, the local node embedding $z_v$ only contains local information of a node $v$. However, $z_v$ can be used to guide non-local aggregation, as distant but informative nodes are likely to have similar node features and local structures. Based on this intuition, we propose the attention-guided sorting to enable the non-local aggregation.

\textbf{Attention-Guided Sorting:} The basic idea of the attention-guided sorting is to learn an ordering of nodes, where distant but informative nodes are put near each other. Specifically, given the local node embedding $z_v$ obtained through the local embedding step, we compute one set of attention scores by
\begin{equation}\label{eqn:attn_sort}
a_v = \textsc{Attend}(c, z_v) \in \mathbb{R}, \  \forall v \in V,
\end{equation}
where $c$ is a calibration vector that is randomly initialized and jointly learned during training~\cite{yang2016hierarchical}. In this attention operator, $c$ serves as the query vector and $z_v$ are the key vectors. In addition, we also treat $z_v$ as the value vectors. However, unlike the attention mechanism introduced in Section~\ref{sec:attn}, we use the attention scores to sort the value vectors instead of computing a weighted sum to aggregating them. Note that originally there is no ordering among nodes in a graph. To be specific, as $a_v$ and $z_v$ have one-to-one correspondence through Eq.~(\ref{eqn:attn_sort}), sorting the attention scores in non-decreasing order into $(a_1, a_2, \ldots, a_n)$ provides an ordering among nodes, where $n = |V|$ is the number of nodes in the graph. The resulting sequence of local node embeddings can be denoted as $(z_1, z_2, \ldots, z_n)$.

The attention process in Eq.~(\ref{eqn:attn_sort}) can be also understood as a projection of local node embeddings onto a 1-dimensional space. The projection depends on the concrete $\textsc{Attend}(\cdot)$ function and the calibration vector $c$. As indicated by its name, the calibration vector $c$ is used to calibrate the 1-dimensional space, in order to push distant but informative nodes close to each other in this space. This goal is fulfilled through the following non-local aggregation step and the training of the calibration vector $c$, as demonstrated below.

\textbf{Non-Local Aggregation:} We point out that, with the attention-guided sorting, the non-local aggregation can be achieved by convolution, the most common local aggregation operator in deep learning. Specifically, given the sorted sequence of local node embeddings $(z_1, z_2, \ldots, z_n)$, we compute
\begin{equation}\label{eqn:conv}
(\hat{z}_1, \hat{z}_2, \ldots, \hat{z}_n) = \textsc{Conv}(z_1, z_2, \ldots, z_n),
\end{equation}
where the $\textsc{Conv}(\cdot)$ function represents a 1D convolution with appropriate padding. Note that $\textsc{Conv}(\cdot)$ can be replaced by a 1D convolutional neural network as long as the number of input and output vectors remains the same.

To see how the $\textsc{Conv}(\cdot)$ function performs non-local aggregation with the attention-guided sorting, we take an example where the $\textsc{Conv}(\cdot)$ function is a 1D convolution of kernel size $2s+1$. In this case, $\hat{z}_i$ is computed from $(z_{i+s}, \ldots, z_{i-s})$, corresponding to the receptive field of the $\textsc{Conv}(\cdot)$ function. As a result, if the attention-guided sorting leads to $(z_{i+s}, \ldots, z_{i-s})$ containing nodes that are distant but informative to $z_i$, the output $\hat{z}_i$ aggregates non-local information. Another view is that we can consider the attention-guided sorting as re-connects nodes in the graph, where $(z_{i+s}, \ldots, z_{i-s})$ can be treated as the 1-hop neighborhood of $z_i$. After the $\textsc{Conv}(\cdot)$ function, $\hat{z}_i$ and $z_i$ are concatenated as the input to a classifier to predict the label of the corresponding node, where both non-local and local dependencies can be captured. In order to enable the end-to-end training of the calibration vector $c$, we modify Eq.~(\ref{eqn:conv}) into
\begin{equation}\label{eqn:conv1}
(\hat{z}_1, \hat{z}_2, \ldots, \hat{z}_n) = \textsc{Conv}(a_1 z_1, a_2 z_2, \ldots, a_n z_n),
\end{equation}
where we multiply the attention score with the corresponding local node embedding. As a result, the calibration vector $c$ receives \textcolor{my_color}{back-propagated} gradients through the attention scores during training.

The remaining question is how to make sure that the attention-guided sorting pushes distant but informative nodes together. We can understand this intuitively by comparing our attention-guided sorting with the attention mechanism that uses learnable query~\cite{yang2016hierarchical}. In such attention mechanism, attention scores obtained by attending learnable query vector to key vectors are used to compute the weighted sum of value vectors. In this case, the learnable query vector can be optimized towards its goal, \emph{i.e.}, assigning appropriate weights to value vectors, since the attention scores serve as weights directly. In our attention-guided sorting, the calibration vector is trained towards its goal, \emph{i.e.}, deriving a good sorting, in an indirect manner. Specifically, the attention scores can be treated as unnormalized weights. At the beginning of training, the calibration vector yields random weights and a correspondingly random sorting. Thus the near nodes might not be informative. Notably, nodes in the same receptive field of the subsequent 1D convolutional neural network tends to have similar weights as all the nodes are already sorted according to the weights. Hence, if the near nodes are not informative, the gradient back-propagated from the classification loss tends to tune the calibration vector to produce better weights and a correspondingly useful sorting. Hence, the calibration vector can be trained to obtain a good sorting effectively.

Also, note that the requirement of non-local aggregation depends on the concrete graphs. In fact, our proposed framework grants GNNs the ability of non-local aggregation but lets the end-to-end training process determine whether to use non-local information. The back-propagation from the supervised loss will tune the calibration vector $c$ effectively and encourage $\hat{z}_i$ to capture useful information that is not encoded by $z_i$. In the case of disassortative graphs, $\hat{z}_i$ usually needs to aggregate information from distant but informative nodes. Hence, the calibration vector $c$ tends to arrange the attention-guided sorting to put distant but informative nodes together, as demonstrated experimentally in Section~\ref{sec:exp_visualize}. On the other hand, nodes within the local neighborhood are usually much more informative than distant nodes in assortative graphs. In this situation, $\hat{z}_i$ may simply perform local aggregation that is similar to GNNs.

\subsection{Time Complexity Analysis}\label{sec:time_complexity}

We perform theoretical analysis of the time complexity of our proposed framework. As discussed in Section~\ref{sec:attn}, using the attention mechanism~\cite{vaswani2017attention,wang2018non,wang2020non} to achieve non-local aggregation requires $O(n^2)$ time for a graph of $n$ nodes. Essentially, the $O(n^2)$ time complexity is due to the fact that the $\textsc{Attend}(\cdot)$ function needs to be computed between every pair of nodes. In particular, the recently proposed Geom-GCN~\cite{pei2020geom} contains a similar non-local aggregation step. For each $v \in V$, Geom-GCN finds the set of nodes from which the Euclidean distance to $v$ is less than a pre-defined number, where the Euclidean distance between every pair of nodes needs to be computed. As the computation of the the Euclidean distance between two nodes can be understood as the $\textsc{Attend}(\cdot)$ function, Geom-GCN has at least $O(n^2)$ time complexity.

In contrast, our proposed non-local aggregation framework requires only $O(n\log n)$ time. To see this, note that the $\textsc{Attend}(\cdot)$ function in Eq.~(\ref{eqn:attn_sort}) only needs to be computed once, instead of iterating it for each node. As a result, computing the attention scores only takes $O(n)$ time. Therefore, the time complexity of sorting, \emph{i.e.} $O(n\log n)$, dominates the total time complexity of our proposed framework. In Section~\ref{sec:exp_efficiency}, we compare the real running time on different datasets among common GNNs, Geom-GCN, and our non-local GNNs as introduced in the next section.

\subsection{Efficient Non-Local Graph Neural Networks}\label{sec:nlgnn}

We apply our non-local aggregation framework to build efficient non-local GNNs. Recall that our proposed framework starts with the local embedding step, followed by the attention-guided sorting and the non-local aggregation step.

In particular, the local embedding step can be implemented by either MLP or common GNNs, such as GCNs~\cite{kipf2016semi} or GATs~\cite{velivckovic2017graph}. MLP extracts the local node embedding only from the node feature vector and excludes the information from nodes within the local neighborhood. This property can be helpful on some disassortative graphs, where nodes within the local neighborhood provide more noises than useful information. On other disassortative graphs, informative nodes locate in both local neighborhood and distant locations. In this case, GNNs are more suitable as the local embedding step. Depending on the disassortative graphs in hand, we build different non-local GNNs with either MLP or GNNs as the local embedding step. In Section~\ref{sec:exp_dis}, we show that these two categories of disassortative graphs can be distinguished through simple experiments, where we apply different non-local GNNs accordingly. Specifically, the number of layers is set to 2 for both MLP and GNNs.

\begin{table*}[t]
	\caption{Statistics of the \textcolor{my_color}{eleven} datasets used in our experiments. The definition of $H(\mathcal{G})$ is provided in Section~\ref{sec:dataset}. $H(\mathcal{G})$ can be used to distinguish assortative and disassortative graph datasets.}
	\label{tab:dataset}
    \vspace{-0.15in}
	\centering
		\resizebox{\textwidth}{!}{
		\begin{tabular}{lcccc|ccccccc}
			\toprule
			& \multicolumn{3}{c}{\textbf{Assortative}} & \multicolumn{6}{c}{\textbf{Disassortative}} \\
			\textbf{Datasets} & \textit{Cora} &\textit{Citeseer} &\textit{Pubmed} & \textit{ogbn-arxiv} & \textit{snap-patents} &\textit{Chameleon} &\textit{Squirrel} &\textit{Actor} &\textit{Cornell} &\textit{Texas} &\textit{Wisconsin} \\
			\midrule
			$H(\mathcal{G})$ &$0.83$ &$0.71$ &$0.79$ & $0.64$ & $0.22$ & $0.25$ & $0.22$ & $0.24$ & $0.11$ & $0.06$ & $0.16$ \\
			Splits & $60/20/20$ & $60/20/20$ & $60/20/20$ & $54/18/28$ & $50/25/25$ & $60/20/20$&$60/20/20$ &$60/20/20$ &$60/20/20$ &$60/20/20$ &$60/20/20$\\
			\#Nodes &$2708$ &$3327$ &$19717$ & $169343$ & $2923922$ & $2277$ & $5201$ & $7600$ & $183$ & $183$ & $251$ \\
			\#Edges &$5429$ &$4732$ &$44338$ & $1166243$  & $13975788$ & $36101$ & $217073$ & $33544$ & $295$ & $309$ & $499$ \\
			\#Features &$1433$ &$3703$ &$500$ & $128$ & $269$ & $2325$ & $2089$ & $931$ & $1703$ & $1703$ & $1703$ \\
			\#Classes &$7$ &$6$ &$3$ & $40$ & $5$  & $5$ & $5$ & $5$ & $5$ & $5$ & $5$ \\
			\bottomrule
		\end{tabular}
		}
\end{table*}

In terms of the attention-guided sorting, we need to specify the $\textsc{Attend}(\cdot)$ function in Eq.~(\ref{eqn:attn_sort}). In order to make it as efficient as possible, we choose it as
\begin{equation}\label{eqn:attn_sort1}
a_v = \textsc{Attend}(c, z_v) = c^T z_v \in \mathbb{R}, \  \forall v \in V,
\end{equation}
where $c$ is part of the training parameters.

With the attention-guided sorting, we can implement the non-local aggregation step through convolution, as explained in Section~\ref{sec:framework}. Specifically, $\textsc{Conv}(\cdot)$ function is set as a 2-layer convolutional neural network composed of two 1D convolutions. The kernel size is \textcolor{my_color}{treated as a hyperparameter}. The activation function is ReLU~\cite{krizhevsky2012imagenet}.

Finally, we use a linear classifier that takes the concatenation of $\hat{z}_i$ and $z_i$ as inputs and makes prediction for the corresponding node. Depending on the local embedding step, we build three efficient non-local GNNs, namely non-local MLP~(NLMLP), non-local GCN~(NLGCN), and non-local GAT~(NLGAT). The models can be end-to-end trained with the classification loss.

\section{Experiments}\label{sec:exp}

\subsection{Datasets}\label{sec:dataset}

We perform experiments on \textcolor{my_color}{seven} disassortative graph datasets~\cite{lim2021new,rozemberczki2019multi,tang2009social,pei2020geom} (\textcolor{my_color}{\textit{snap-patents}}, \textit{Chameleon}, \textit{Squirrel}, \textit{Actor}, \textit{Cornell}, \textit{Texas}, \textit{Wisconsin}) and \textcolor{my_color}{four} assortative graph datasets~\cite{sen2008collective,hu2020open} (\textit{Cora}, \textit{Citeseer}, \textit{Pubmed}, \textcolor{my_color}{\textit{ogbn-arxiv}}). These datasets are commonly used to evaluate GNNs on node classification tasks~\cite{kipf2016semi,velivckovic2017graph,gao2018large,pei2020geom}. We provide detailed descriptions of disassortative graph datasets below.

\textcolor{my_color}{\textit{snap-patents} is an extremely large patent graph~\cite{leskovec2014snap} where nodes are patents and edges denote citation relationships. Node features are extracted from patent metadata. Each node is labeled by the year when the corresponding patent war granted.}

\textit{Chameleon} and \textit{Squirrel} are Wikipedia networks~\cite{rozemberczki2019multi} where nodes represent web pages from Wikipedia and edges indicate mutual links between pages. Node feature vectors are bag-of-word representations of informative nouns in the corresponding pages. Each node is labeled with one of five classes according to the number of average monthly traffic of the web page.

\textit{Actor} is an actor co-occurrence network, where nodes denote actors and edges indicate co-occurrence on the same web page from Wikipedia. It is extracted from the film-director-actor-writer network in~\cite{tang2009social}. Node feature vectors are bag-of-word representations of keywords in the actors' Wikipedia pages. Each node is labeled with one of five classes according to the topic of the actor’s Wikipedia page.

\textit{Cornell}, \textit{Texas}, and \textit{Wisconsin} come from the WebKB dataset collected by Carnegie Mellon University. Nodes represent web pages and edges denote hyperlinks between them. Node feature vectors are bag-of-word representations of the corresponding web pages. Each node is labeled as student, project, course, staff, or faculty.

In order to distinguish assortative and disassortative graph datasets,~\cite{pei2020geom} proposes a metric to measure the homophily of a graph $\mathcal{G}$, defined as
\begin{equation}
H(\mathcal{G}) = \frac{1}{|V|}\sum_{v \in V}\frac{|\{u\text{: }u \in \mathcal{N}_1(v) \text{ and } l(u)=l(v)\}|}{|\mathcal{N}_1(v)|},
\end{equation}
where $|\{u\text{: }u \in \mathcal{N}_1(v) \text{ and } l(u)=l(v)\}|$ denotes the number of $v$'s directly connected nodes who have the same label as $v$ and $|\mathcal{N}_1(v)|$ represents the number of $v$'s directly connected nodes. Intuitively, a large $H(\mathcal{G})$ indicates an assortative graph, and vice versa. The $H(\mathcal{G})$ and other statistics for all datasets are summarized in Table~\ref{tab:dataset}.

\subsection{Baselines}\label{sec:baselines}

We compare our proposed non-local MLP~(NLMLP), non-local GCN~(NLGCN), and non-local GAT~(NLGAT) with various baselines: (1) MLP is the simplest deep learning model. It makes prediction solely based on the node feature vectors, without aggregating any local or non-local information. (2) GCN~\cite{kipf2016semi} and GAT~\cite{velivckovic2017graph} are the most common GNNs. As introduced in Section~\ref{sec:gnn}, they only perform local aggregation. (3) Geom-GCN~\cite{pei2020geom} is a recently proposed GNN that can capture long-range dependencies. Geom-GCN requires the use of different node embedding methods, such as Isomap~\cite{tenenbaum2000global}, Poincare~\cite{nickel2017poincare}, and struc2vec~\cite{ribeiro2017struc2vec}. We simply report the best results from~\cite{pei2020geom} for Geom-GCN and the following two variants without specifying the node embedding method. (4) Geom-GCN-g~\cite{pei2020geom} is a variant of Geom-GCN that performs local aggregation only. It is similar to common GNNs. (5) Geom-GCN-s~\cite{pei2020geom} is a variant of Geom-GCN that does not force local aggregation. The designed functionality is similar to our NLMLP. \textcolor{my_color}{(6) In addition, we also consider several recently developed methods on disassortative graphs as baselines. They are H$_2$GCN~\cite{zhu2020beyond}, SimP-GCN~\cite{jin2021node}, FAGCN~\cite{bo2021beyond}, and CPGCN~\cite{zhu2021graph}. \textcolor{my_color}{Since H$_2$GCN and CPGCN has multiple variants, we report its best result on each dataset.}} 

We implement MLP, GCN, GAT, and our methods using Pytorch~\cite{paszke2017automatic} and Pytorch Geometric~\cite{Fey2019pyg}. As has been discussed\footnote{\url{https://openreview.net/forum?id=S1e2agrFvS&noteId=8tGKV1oSzCr}}, in fair settings, the results of GCN and GAT differ from those in~\cite{pei2020geom}.

\begin{table*}[t!]
	\caption{\textcolor{my_color}{Comparisons between MLP and common GNNs in terms of the best validation accuracy. These analytical experiments are used to determine the two categories of disassortative graph datasets, as introduced in Section~\ref{sec:exp_dis}.}}
	\label{tab:result_dis}
   \vspace{-0.15in}
	\centering
	\resizebox{\textwidth}{!}{
		\begin{tabular}{lcccc|ccccccc}
			\toprule
			& \multicolumn{3}{c}{\textbf{Assortative}} & \multicolumn{6}{c}{\textbf{Disassortative}} \\
			\textbf{Datasets} & \textit{Cora} &\textit{Citeseer} &\textit{Pubmed} &\textit{ogbn-arxiv} & \textit{snap-patents}&\textit{Chameleon} &\textit{Squirrel} &\textit{Actor} &\textit{Cornell} &\textit{Texas} &\textit{Wisconsin} \\
			\midrule
			MLP &$77.2 \scriptstyle{{\light{\pm1.6}}}$ &$74.4 \scriptstyle{{\light{\pm1.4}}}$ &$87.4 \scriptstyle{{\light{\pm0.5}}}$ & $56.6 \scriptstyle{{\light{\pm0.1}}}$ & $31.2 \scriptstyle{{\light{\pm0.1}}}$ & $47.8 \scriptstyle{{\light{\pm1.7}}}$ & $33.3 \scriptstyle{{\light{\pm0.8}}}$ & $\textbf{35.9} \scriptstyle{{\light{\pm0.9}}}$ & $\textbf{83.1} \scriptstyle{{\light{\pm3.3}}}$ & $\textbf{81.5} \scriptstyle{{\light{\pm3.9}}}$ & $\textbf{86.7}\scriptstyle{{\light{\pm2.4}}}$  \\
			GCN &$88.8 \scriptstyle{{\light{\pm1.0}}}$ &$\textbf{78.1} \scriptstyle{{\light{\pm1.3}}}$ &$\textbf{88.7} \scriptstyle{{\light{\pm0.5}}}$ & $\textbf{71.5} \scriptstyle{{\light{\pm0.1}}}$ & $\textbf{46.2} \scriptstyle{{\light{\pm0.6}}}$ & $\textbf{67.7} \scriptstyle{{\light{\pm1.3}}}$ & $\textbf{54.2}\scriptstyle{{\light{\pm1.1}}}$ & $32.2 \scriptstyle{{\light{\pm0.8}}}$ & $58.5 \scriptstyle{{\light{\pm2.7}}}$ & $68.1 \scriptstyle{{\light{\pm1.6}}}$ & $62.3\scriptstyle{{\light{\pm3.5}}}$  \\
			GAT& $\textbf{89.8} \scriptstyle{{\light{\pm1.4}}}$ &$77.6 \scriptstyle{{\light{\pm1.3}}}$ &$87.7 \scriptstyle{{\light{\pm0.4}}}$ & $66.5 \scriptstyle{{\light{\pm2.5}}}$ & $45.6 \scriptstyle{{\light{\pm0.5}}}$  & $66.4 \scriptstyle{{\light{\pm1.3}}}$ & $51.8\scriptstyle{{\light{\pm2.6}}}$ & $30.6 \scriptstyle{{\light{\pm0.6}}}$ & $61.9 \scriptstyle{{\light{\pm1.6}}}$ & $63.2 \scriptstyle{{\light{\pm4.7}}}$ & $64.3\scriptstyle{{\light{\pm3.5}}}$  \\
			\bottomrule
		\end{tabular}
		}
\end{table*}

\textcolor{my_color}{The experiments on \textit{snap-patents} are repeatedly conducted by $5$ times, following~\cite{lim2021new}, and experiments on other datasets are repeatedly performed 10 times. The average test accuracy over runs are reported unless stated specifically.} Testing is performed when validation accuracy achieves maximum on each run. Apart from the details specified in Section~\ref{sec:nlgnn}, \textcolor{my_color}{we tune the following hyperparameters on validation set for our proposed models: (1) the number of hidden unit $\in$ \{16, 48, 96, 128, 256\}, (2) the kernel size of convolution $\in$ \{3, 5, 7\}, (3) dropout rate $\in$ \{0, 0.5, 0.8\}, (4) weight decay $\in$ \{0, 5e-4, 5e-5, 5e-6\}, and (5) learning rate $\in$ \{0.001, 0.01, 0.05\}.}

\subsection{Analysis of Disassortative Graph Datasets}\label{sec:exp_dis}

As discussed in Section~\ref{sec:nlgnn}, the disassortative graph datasets can be divided into two categories. Nodes within the local neighborhood provide more noises than useful information in disassortative graphs belonging to the first category. Therefore, local aggregation should be avoided in models on such disassortative graphs. As for the second category, informative nodes locate in both local neighborhood and distant locations. Intuitively, a graph with lower $H(\mathcal{G})$ is more likely to be in the first category. However, it is not an accurate way to determine the two categories.

Knowing the exact category of a disassortative graph is crucial, as we need to apply non-local GNNs accordingly. As analyzed above, the key difference lies in whether the local aggregation is useful. Hence, we can distinguish two categories of disassortative graphs by comparing the performance between MLP and common GNNs (GCN, GAT) on each of the \textcolor{my_color}{seven} disassortative graph datasets. \textcolor{my_color}{To be rigorous, we compare the average of the best validation accuracy over runs in Table~\ref{tab:result_dis}.}

We can see that \textit{Actor}, \textit{Cornell}, \textit{Texas}, and \textit{Wisconsin} fall into the first category, while \textcolor{my_color}{\textit{snap-patents}}, \textit{Chameleon}, and \textit{Squirrel} belong to the second category. We add the performance on assortative graph datasets for reference, where the local aggregation is effective so that GNNs tend to outperform MLP.

\subsection{Comparisons with Baselines}\label{sec:results}

\begin{table}[t]
	\caption{Comparisons between our NLMLP and strong baselines on the four disassortative graph datasets belonging to the first category as defined in Section~\ref{sec:exp_dis}.}
	\label{tab:result_mlp}
	\vspace{-0.15in}
	\centering
	\resizebox{\columnwidth}{!}{
	\begin{tabular}{lcccc}
		\toprule
		\textbf{Datasets} &\textit{Actor} &\textit{Cornell} &\textit{Texas} &\textit{Wisconsin} \\
		\midrule
		MLP &$35.1 \scriptstyle{{\light{\pm0.8}}}$ & $81.6 \scriptstyle{{\light{\pm6.3}}}$ & $81.3 \scriptstyle{{\light{\pm7.1}}}$ & $84.9\scriptstyle{{\light{\pm5.3}}}$  \\
		\midrule
		Geom-GCN & $31.6$ & $60.8$ & $67.6$ & $64.1$ \\
		Geom-GCN-s & $34.6$ & $75.4$ & $73.5$ & $80.4$ \\
		\midrule
		H$_2$GCN & $35.9 \scriptstyle{{\light{\pm1.0}}}$ & $82.2 \scriptstyle{{\light{\pm6.0}}}$ & $84.9 \scriptstyle{{\light{\pm6.8}}}$ & $86.7\scriptstyle{{\light{\pm4.7}}}$  \\
		SimP-GCN & $36.2$ & $84.1$ & $81.6$ & $85.5$  \\
		\midrule
		\textbf{NLMLP} (ours) & $\textbf{37.9} \scriptstyle{{\light{\pm1.3}}}$ & $\textbf{84.9} \scriptstyle{{\light{\pm5.7}}}$ & $\textbf{85.4} \scriptstyle{{\light{\pm3.8}}}$ & $\textbf{87.3}\scriptstyle{{\light{\pm4.3}}}$  \\
		\bottomrule
	\end{tabular}
}
\end{table}

According to the insights from Section~\ref{sec:exp_dis}, we apply different non-local GNNs according to the category of disassortative graph datasets, and make comparisons with corresponding baselines. In our experiments, we focus on comparing the model performance on disassortative graph datasets, in order to demonstrate the effectiveness of our non-local aggregation framework. The performances on assortative graph datasets are also included in Table~\ref{tab:all_results} for reference, indicating that the proposed framework will not hurt the performance when non-local aggregation is not strongly desired.

Specifically, we employ NLMLP on \textit{Actor}, \textit{Cornell}, \textit{Texas}, and \textit{Wisconsin}. The corresponding baselines are MLP, Geom-GCN, and Geom-GCN-s, as Table~\ref{tab:result_dis} has shown that GCN and GAT perform much worse than MLP on these datasets. And Geom-GCN-g is similar to GCN and has worse performance than Geom-GCN-s, which is shown in Table~\ref{tab:all_results}. \textcolor{my_color}{We also include H$_2$GCN and SimP-GCN as baselines since they also conduct experiments on these datasets.} The comparison results are reported in Table~\ref{tab:result_mlp}. We find that MLP consistently outperforms Geom-GCN-s by large margins. In particular, although Geom-GCN-s does not explicitly perform local aggregation, it is still outperformed by MLP. A possible explanation is that Geom-GCN-s uses pre-trained node embeddings, which aggregates information from the local neighborhood implicitly. In contrast, our NLMLP is built upon MLP with the proposed non-local aggregation framework, which excludes the local noises and collects useful information from non-local informative nodes. The NLMLP sets the new state-of-the-art performance on these disassortative graph datasets.

\begin{table}
	\caption{Comparisons between our NLGCN, NLGAT and strong baselines on the \textcolor{my_color}{three} disassortative graph datasets belonging to the second category as defined in Section~\ref{sec:exp_dis}.}
	\label{tab:result_gnn}
	\centering
	\vspace{-0.15in}
	\begin{tabular}{lccc}
		\toprule
		\textbf{Datasets} & \textit{snap-patents} &\textit{Chameleon} &\textit{Squirrel} \\
		\midrule
		GCN & $45.7 \scriptstyle{{\light{\pm0.0}}}$ & $67.6 \scriptstyle{{\light{\pm2.4}}}$ & $54.9\scriptstyle{{\light{\pm1.9}}}$ \\
		GAT & $45.4 \scriptstyle{{\light{\pm0.4}}}$ & $65.0 \scriptstyle{{\light{\pm3.7}}}$ & $51.3\scriptstyle{{\light{\pm2.5}}}$\\
		\midrule
		Geom-GCN & - & $60.9$ & $38.1$ \\
		Geom-GCN-g & - & $68.0$ & $46.0$ \\
		\midrule
		H$_2$GCN & - & $59.4 \scriptstyle{{\light{\pm2.0}}}$ & $37.9 \scriptstyle{{\light{\pm2.0}}}$ \\
		FAGCN & - & $61.7$ & $39.7$ \\
		\midrule
		\textbf{NLGCN} (ours) & $\textbf{50.8} \scriptstyle{{\light{\pm0.4}}}$ & $\textbf{70.1} \scriptstyle{{\light{\pm2.9}}}$ & $\textbf{59.0}\scriptstyle{{\light{\pm1.2}}}$\\
		\textbf{NLGAT} (ours) & $48.3 \scriptstyle{{\light{\pm0.2}}}$ & $65.7 \scriptstyle{{\light{\pm1.4}}}$ & $56.8\scriptstyle{{\light{\pm2.5}}}$ \\
		\bottomrule
	\end{tabular}
\end{table}

On \textcolor{my_color}{\textit{snap-patents}}, \textit{Chameleon}, and \textit{Squirrel} that belong to the second category of disassortative graph datasets, we apply NLGCN and NLGAT accordingly. The baselines are GCN, GAT, Geom-GCN, Geom-GCN-g, \textcolor{my_color}{H$_2$GCN, FAGCN, and CPGCN}. On these datasets, these baselines that explicitly perform local aggregation show advantages over MLP and Geom-GCN-s, as shown in Table~\ref{tab:result_dis} and~\ref{tab:all_results}. As shown in Table~\ref{tab:result_gnn}, our proposed NLGCN achieves the best performance on \textcolor{my_color}{all three} datasets. \textcolor{my_color}{Note that CPGCN uses different splits as other baselines. For comparing with CPGCN, we additionally run our NLGCN using the same splits as CPGCN. On \textit{Chameleon} and \textit{Squirrel}, our NLGCN achieves better test accuracies of $41.3$ and $60.0$ respectively, compared to $37.0$ and $56.9$ of CPGCN.} In addition, it is worth noting that our NLGCN and NLGAT are built upon GCN and GAT, respectively. They show improvements over their counterparts, which indicates that the advantages of our proposed non-local aggregation framework are general for common GNNs.

\begin{figure*}[th]
	\centering
	\includegraphics[width=0.9\textwidth]{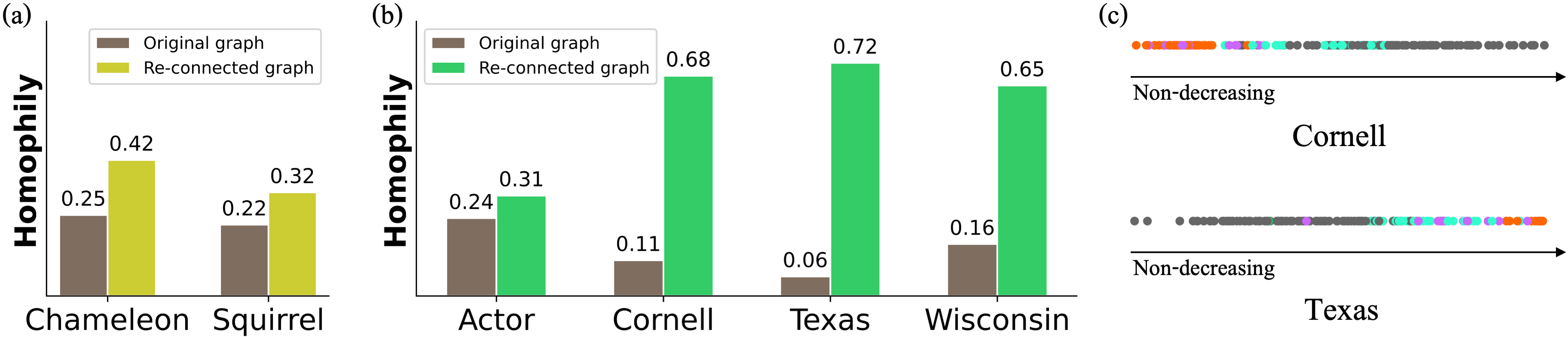}
	\vspace{-0.15in}
	\caption{(a) Comparisons of the homophily between the original graph and the re-connected graph given by our NLGCN on \textit{Chameleon} and \textit{Squirrel}. (b) Comparisons of the homophily between the original graph and the re-connected graph given by our NLMLP on \textit{Actor}, \textit{Cornell}, \textit{Texas}, and \textit{Wisconsin}. (c) Visualization of sorted node sequence after the attention-guided sorting for \textit{Cornell} and \textit{Texas}. The colors denote node labels. Details are explained in Section~\ref{sec:exp_visualize}.}
	\label{fig:visualization}
\end{figure*}

\begin{table*}[ht]
	\caption{Comparisons between our NLMLP, NLGCN, NLGAT and baselines on all the $11$ datasets. \textcolor{my_color}{$^\dagger$We can only use smaller hidden dimensions for GAT and NLGAT on \textit{ogbn-arxiv} and \textit{snap-patents} due to the huge memory requirement. $^\ddagger$Uses a GPU with $48$ GB of memory.}}
	\label{tab:all_results}
	\centering
	\vspace{-0.15in}
	\resizebox{\textwidth}{!}{
		\begin{tabular}{lcccc|ccccccc}
			\toprule
			& \multicolumn{3}{c}{\textbf{Assortative}} & \multicolumn{6}{c}{\textbf{Disassortative}} \\
			\textbf{Datasets} & \textit{Cora} &\textit{Citeseer} &\textit{Pubmed} &\textit{ogbn-arxiv} & \textit{snap-patents} &\textit{Chameleon} &\textit{Squirrel} &\textit{Actor} &\textit{Cornell} &\textit{Texas} &\textit{Wisconsin} \\
			\midrule
			MLP &$76.5 \scriptstyle{{\light{\pm1.3}}}$ &$73.6 \scriptstyle{{\light{\pm1.9}}}$ &$87.5 \scriptstyle{{\light{\pm0.4}}}$ & $54.1 \scriptstyle{{\light{\pm0.1}}}$ & $31.3 \scriptstyle{{\light{\pm0.1}}}$ & $48.5 \scriptstyle{{\light{\pm3.0}}}$ & $31.5 \scriptstyle{{\light{\pm1.4}}}$ & $35.1 \scriptstyle{{\light{\pm0.8}}}$ & $81.6 \scriptstyle{{\light{\pm6.3}}}$ & $81.3 \scriptstyle{{\light{\pm7.1}}}$ & $84.9\scriptstyle{{\light{\pm5.3}}}$  \\
			GCN &$88.2 \scriptstyle{{\light{\pm1.2}}}$ &$75.7 \scriptstyle{{\light{\pm1.3}}}$ &$88.4 \scriptstyle{{\light{\pm0.6}}}$ & $70.4 \scriptstyle{{\light{\pm0.2}}}$  & $45.7 \scriptstyle{{\light{\pm0.0}}}$ & $67.6 \scriptstyle{{\light{\pm2.4}}}$ & $54.9\scriptstyle{{\light{\pm1.9}}}$ & $30.3 \scriptstyle{{\light{\pm1.6}}}$ & $54.2 \scriptstyle{{\light{\pm7.3}}}$ & $61.1 \scriptstyle{{\light{\pm7.0}}}$ & $59.6\scriptstyle{{\light{\pm4.5}}}$  \\
			GAT& $88.4 \scriptstyle{{\light{\pm1.4}}}$ &$76.1 \scriptstyle{{\light{\pm1.0}}}$ &$87.0 \scriptstyle{{\light{\pm0.3}}}$ & $65.0 \scriptstyle{{\light{\pm2.9}}}^\dagger$ & $45.4 \scriptstyle{{\light{\pm0.4}}}^\dagger$ & $65.0 \scriptstyle{{\light{\pm3.7}}}$ & $51.3\scriptstyle{{\light{\pm2.5}}}$ & $29.4 \scriptstyle{{\light{\pm1.2}}}$ & $56.3 \scriptstyle{{\light{\pm4.3}}}$ & $57.9 \scriptstyle{{\light{\pm6.1}}}$ & $57.8\scriptstyle{{\light{\pm4.3}}}$  \\
			\midrule
			Geom-GCN &$85.3$ &$78.0$ &$90.1$ & - & - & $60.9$ & $38.1$ & $31.6$ & $60.8$ & $67.6$ & $64.1$ \\
			Geom-GCN-g &$87.0$ &$\textbf{80.6}$ &$\textbf{90.7}$ & - & - & $68.0$ & $46.0$ & $32.0$ & $67.0$ & $73.1$ & $69.4$  \\
			Geom-GCN-s &$73.3$ &$72.2$ &$87.0$ & - & - & $61.6$ & $38.0$ & $34.6$ & $75.4$ & $73.5$ & $80.4$ \\
			\midrule
			\textbf{NLMLP} (ours) & $76.9 \scriptstyle{{\light{\pm1.8}}}$ &$73.4 \scriptstyle{{\light{\pm1.9}}}$ &$88.2  \scriptstyle{{\light{\pm0.5}}}$ & $54.0 \scriptstyle{{\light{\pm0.3}}}$ & $32.1 \scriptstyle{{\light{\pm0.1}}}^\ddagger$& $50.7 \scriptstyle{{\light{\pm2.2}}}$ & $33.7\scriptstyle{{\light{\pm1.5}}}$ & $\textbf{37.9} \scriptstyle{{\light{\pm1.3}}}$ & $\textbf{84.9} \scriptstyle{{\light{\pm5.7}}}$ & $\textbf{85.4} \scriptstyle{{\light{\pm3.8}}}$ & $\textbf{87.3}\scriptstyle{{\light{\pm4.3}}}$  \\
			\textbf{NLGCN} (ours) &$88.1 \scriptstyle{{\light{\pm1.0}}}$ &$75.2 \scriptstyle{{\light{\pm1.4}}}$ &$89.0 \scriptstyle{{\light{\pm0.5}}}$ &$\textbf{70.6} \scriptstyle{{\light{\pm0.3}}}$ & $\textbf{50.8} \scriptstyle{{\light{\pm0.4}}}^\ddagger$ & $\textbf{70.1} \scriptstyle{{\light{\pm2.9}}}$ & $\textbf{59.0}\scriptstyle{{\light{\pm1.2}}}$ & $31.6 \scriptstyle{{\light{\pm1.0}}}$ & $57.6 \scriptstyle{{\light{\pm5.5}}}$ & $65.5 \scriptstyle{{\light{\pm6.6}}}$ & $60.2\scriptstyle{{\light{\pm5.3}}}$  \\
			\textbf{NLGAT} (ours) &$\textbf{88.5} \scriptstyle{{\light{\pm1.8}}}$ &$76.2 \scriptstyle{{\light{\pm1.6}}}$ &$88.2 \scriptstyle{{\light{\pm0.3}}}$ & $66.7 \scriptstyle{{\light{\pm2.4}}}^\dagger$ & $48.3 \scriptstyle{{\light{\pm0.2}}}^{\dagger\ddagger}$ & $65.7 \scriptstyle{{\light{\pm1.4}}}$ & $56.8\scriptstyle{{\light{\pm2.5}}}$ & $29.5 \scriptstyle{{\light{\pm1.3}}}$ & $54.7 \scriptstyle{{\light{\pm7.6}}}$ & $62.6 \scriptstyle{{\light{\pm7.1}}}$ & $56.9\scriptstyle{{\light{\pm7.3}}}$  \\
			\bottomrule
		\end{tabular}
    	}
\end{table*}

We summarize the results on all datasets in Table~\ref{tab:all_results} for reference.

\subsection{Analysis of the Attention-Guided Sorting}\label{sec:exp_visualize}

We analyze the results of the attention-guided sorting in our proposed framework, in order to show that our non-local GNNs indeed perform non-local aggregation.

Suppose the attention-guided sorting leads to the sorted sequence $(z_1, z_2, \ldots, z_n)$, which goes through a convolution or CNN into $(\hat{z}_1, \hat{z}_2, \ldots, \hat{z}_n)$. As discussed in Section~\ref{sec:framework}, we can consider the sequence $(z_1, z_2, \ldots, z_n)$ as a re-connected graph $\mathcal{\hat{G}}$, where we treat nodes within the receptive field of $\hat{z}_i$ as directly connected to $z_i$, \emph{i.e.} $z_i$'s 1-hop neighborhood. The information within this new 1-hop neighborhood will be aggregated. If our non-local GNNs indeed perform non-local aggregation, the homophily of the re-connected graph should be larger than the original graph. Therefore, we compute $H(\mathcal{\hat{G}})$ for several datasets to verify this statement. Following Section~\ref{sec:results}, we apply NLMLP on \textit{Actor}, \textit{Cornell}, \textit{Texas}, and \textit{Wisconsin} and NLGCN on \textit{Chameleon} and \textit{Squirrel}.

Figure~\ref{fig:visualization} compares $H(\mathcal{\hat{G}})$ with $H(\mathcal{G})$. We can observe that $H(\mathcal{\hat{G}})$ is much larger than $H(\mathcal{G})$, indicating that distant but informative nodes are near each other in the re-connected graph $\mathcal{\hat{G}}$. We also provide the visualizations of the sorted sequence for \textit{Cornell} and \textit{Texas}. We can see that nodes with the same label tend to be clustered together. These facts indicate that our non-local GNNs indeed perform non-local aggregation with the attention-guided sorting.

\subsection{Efficiency Comparisons}\label{sec:exp_efficiency}

As analyzed in Section~\ref{sec:time_complexity}, our proposed non-local aggregation framework is more efficient than previous methods based on the original attention mechanism, such as Geom-GCN~\cite{pei2020geom}. Concretely, our method requires only $O(n\log n)$ computation time in contrast to $O(n^2)$. In this section, we compare the real running time to verify our analysis. Specifically, we compare NLGCN with Geom-GCN as well as GCN and GAT. For Geom-GCN, we use the code provided in~\cite{pei2020geom}. Each model is trained for 500 epochs on each dataset and the average training time per epoch is reported. 

\begin{table}
	\caption{Comparisons in terms of real running time (\textit{milliseconds}). \textcolor{my_color}{$^\dagger$We have memory issue if we use the same dimension for GAT as GCN and NLGCN on \textit{snap-patents}.}}
	\label{tab:result_speed}
	\vspace{-0.15in}
	\centering
	\resizebox{\columnwidth}{!}{
	\begin{tabular}{lccc}
		\toprule
		& \textit{snap-patents} &\textit{Chameleon} &\textit{Squirrel} \\
		\midrule
		GCN & $2193.0 \ (1.0 \times)$ & $22.2 \ (1.0 \times)$ & $14.3 \ (1.0 \times)$ \\
		GAT & OOM$^\dagger$ & $33.2 \ (1.5 \times)$ & $163.3 \ (11.4 \times)$ \\
		Geom-GCN & - & $3615.0 \ (163.1 \times)$ & $10430.0 \ (727.3 \times)$ \\
		\midrule
		\textbf{NLGCN} (ours) & $2437.4 \ (1.1 \times)$ & $26.3 \ (1.2 \times)$ & $39.6 \ (2.8 \times)$ \\
		\bottomrule
	\end{tabular}
}
\end{table}

The results are shown in Table~\ref{tab:result_speed}. Although our NLGCN is built upon GCN, it is just slightly slower than GCN and faster than GAT, showing the efficiency of our non-local aggregation framework. On the other hand, Geom-GCN is significantly slower due to the fact that it has $O(n^2)$ time complexity.

\section{Conclusion}\label{sec:concl}

In this work, we propose a simple yet effective non-local aggregation framework for GNNs. The core of the framework is an efficient attention-guided sorting, which enables non-local aggregation through convolution. The proposed framework can be easily used to build non-local GNNs with low computational costs. We perform thorough experiments on node classification tasks to evaluate our proposed method. In particular, we experimentally analyze existing disassortative graph datasets and apply different non-local GNNs accordingly. The results show that our non-local GNNs significantly outperform previous state-of-the-art methods on all benchmark datasets of disassortative graphs, in terms of both accuracy and speed.


%



\ifCLASSOPTIONcompsoc
  \section*{Acknowledgments}
\else
  \section*{Acknowledgment}
\fi

This work was supported in part by National Science Foundation grants IIS-1908198 and DBI-1922969.





\bibliographystyle{IEEEtran}
\bibliography{my_reference}

\begin{thebibliography}{10}
\providecommand{\url}[1]{#1}
\csname url@samestyle\endcsname
\providecommand{\newblock}{\relax}
\providecommand{\bibinfo}[2]{#2}
\providecommand{\BIBentrySTDinterwordspacing}{\spaceskip=0pt\relax}
\providecommand{\BIBentryALTinterwordstretchfactor}{4}
\providecommand{\BIBentryALTinterwordspacing}{\spaceskip=\fontdimen2\font plus
\BIBentryALTinterwordstretchfactor\fontdimen3\font minus
  \fontdimen4\font\relax}
\providecommand{\BIBforeignlanguage}[2]{{%
\expandafter\ifx\csname l@#1\endcsname\relax
\typeout{** WARNING: IEEEtran.bst: No hyphenation pattern has been}%
\typeout{** loaded for the language `#1'. Using the pattern for}%
\typeout{** the default language instead.}%
\else
\language=\csname l@#1\endcsname
\fi
#2}}
\providecommand{\BIBdecl}{\relax}
\BIBdecl

\bibitem{bronstein2017geometric}
M.~M. Bronstein, J.~Bruna, Y.~LeCun, A.~Szlam, and P.~Vandergheynst,
  ``Geometric deep learning: going beyond euclidean data,'' \emph{IEEE Signal
  Processing Magazine}, vol.~34, no.~4, pp. 18--42, 2017.

\bibitem{gilmer2017neural}
J.~Gilmer, S.~S. Schoenholz, P.~F. Riley, O.~Vinyals, and G.~E. Dahl, ``Neural
  message passing for quantum chemistry,'' in \emph{Proceedings of the 34th
  international conference on machine learning}, 2017, pp. 1263--1272.

\bibitem{battaglia2018relational}
P.~W. Battaglia, J.~B. Hamrick, V.~Bapst, A.~Sanchez-Gonzalez, V.~Zambaldi,
  M.~Malinowski, A.~Tacchetti, D.~Raposo, A.~Santoro, R.~Faulkner
  \emph{et~al.}, ``Relational inductive biases, deep learning, and graph
  networks,'' \emph{arXiv preprint arXiv:1806.01261}, 2018.

\bibitem{zhou2018graph}
J.~Zhou, G.~Cui, Z.~Zhang, C.~Yang, Z.~Liu, L.~Wang, C.~Li, and M.~Sun, ``Graph
  neural networks: A review of methods and applications,'' \emph{arXiv preprint
  arXiv:1812.08434}, 2018.

\bibitem{zhang2018graph}
S.~Zhang, H.~Tong, J.~Xu, and R.~Maciejewski, ``Graph convolutional networks:
  Algorithms, applications and open challenges,'' in \emph{International
  Conference on Computational Social Networks}.\hskip 1em plus 0.5em minus
  0.4em\relax Springer, 2018, pp. 79--91.

\bibitem{wu2019comprehensive}
Z.~Wu, S.~Pan, F.~Chen, G.~Long, C.~Zhang, and P.~S. Yu, ``A comprehensive
  survey on graph neural networks,'' \emph{arXiv preprint arXiv:1901.00596},
  2019.

\bibitem{zhang2020deep}
Z.~Zhang, P.~Cui, and W.~Zhu, ``Deep learning on graphs: A survey,'' \emph{IEEE
  Transactions on Knowledge and Data Engineering}, 2020.

\bibitem{kipf2016semi}
T.~N. Kipf and M.~Welling, ``Semi-supervised classification with graph
  convolutional networks,'' in \emph{International Conference on Learning
  Representations}, 2017.

\bibitem{zhang2018link}
M.~Zhang and Y.~Chen, ``Link prediction based on graph neural networks,'' in
  \emph{Advances in Neural Information Processing Systems}, 2018, pp.
  5165--5175.

\bibitem{ying2018hierarchical}
Z.~Ying, J.~You, C.~Morris, X.~Ren, W.~Hamilton, and J.~Leskovec,
  ``Hierarchical graph representation learning with differentiable pooling,''
  in \emph{Advances in neural information processing systems}, 2018, pp.
  4800--4810.

\bibitem{zhang2018end}
M.~Zhang, Z.~Cui, M.~Neumann, and Y.~Chen, ``An end-to-end deep learning
  architecture for graph classification,'' in \emph{Thirty-Second AAAI
  Conference on Artificial Intelligence}, 2018.

\bibitem{gao2019graph}
H.~Gao and S.~Ji, ``Graph {U}-{N}ets,'' in \emph{International Conference on
  Machine Learning}, 2019, pp. 2083--2092.

\bibitem{ma2019graph}
Y.~Ma, S.~Wang, C.~C. Aggarwal, and J.~Tang, ``Graph convolutional networks
  with eigenpooling,'' in \emph{Proceedings of the 25th ACM SIGKDD
  International Conference on Knowledge Discovery \& Data Mining}, 2019, pp.
  723--731.

\bibitem{lee2019self}
J.~Lee, I.~Lee, and J.~Kang, ``Self-attention graph pooling,'' in
  \emph{International Conference on Machine Learning}, 2019, pp. 3734--3743.

\bibitem{Yuan2020StructPool:}
H.~Yuan and S.~Ji, ``Structpool: Structured graph pooling via conditional
  random fields,'' in \emph{International Conference on Learning
  Representations}, 2020.

\bibitem{bianchi2020spectral}
F.~M. Bianchi, D.~Grattarola, and C.~Alippi, ``Spectral clustering with graph
  neural networks for graph pooling,'' in \emph{International Conference on
  Machine Learning}.\hskip 1em plus 0.5em minus 0.4em\relax PMLR, 2020, pp.
  874--883.

\bibitem{yanardag2015deep}
P.~Yanardag and S.~Vishwanathan, ``Deep graph kernels,'' in \emph{Proceedings
  of the 21th ACM SIGKDD International Conference on Knowledge Discovery and
  Data Mining}.\hskip 1em plus 0.5em minus 0.4em\relax ACM, 2015, pp.
  1365--1374.

\bibitem{defferrard2016convolutional}
M.~Defferrard, X.~Bresson, and P.~Vandergheynst, ``Convolutional neural
  networks on graphs with fast localized spectral filtering,'' in
  \emph{Advances in Neural Information Processing Systems}, 2016, pp.
  3844--3852.

\bibitem{hamilton2017inductive}
W.~Hamilton, Z.~Ying, and J.~Leskovec, ``Inductive representation learning on
  large graphs,'' in \emph{Advances in Neural Information Processing Systems},
  2017, pp. 1024--1034.

\bibitem{velivckovic2017graph}
P.~Veli{\v{c}}kovi{\'c}, G.~Cucurull, A.~Casanova, A.~Romero, P.~Lio, and
  Y.~Bengio, ``Graph attention networks,'' in \emph{International Conference on
  Learning Representation}, 2018.

\bibitem{gao2018large}
H.~Gao, Z.~Wang, and S.~Ji, ``Large-scale learnable graph convolutional
  networks,'' in \emph{Proceedings of the 24th ACM SIGKDD International
  Conference on Knowledge Discovery \& Data Mining}, 2018, pp. 1416--1424.

\bibitem{xu2018powerful}
K.~Xu, W.~Hu, J.~Leskovec, and S.~Jegelka, ``How powerful are graph neural
  networks?'' in \emph{International Conference on Learning Representations},
  2019.

\bibitem{errica2019fair}
F.~Errica, M.~Podda, D.~Bacciu, and A.~Micheli, ``A fair comparison of graph
  neural networks for graph classification,'' in \emph{International Conference
  on Learning Representations}, 2020.

\bibitem{lecun1998gradient}
Y.~LeCun, L.~Bottou, Y.~Bengio, and P.~Haffner, ``Gradient-based learning
  applied to document recognition,'' \emph{Proceedings of the IEEE}, vol.~86,
  no.~11, pp. 2278--2324, 1998.

\bibitem{wang2018non}
X.~Wang, R.~Girshick, A.~Gupta, and K.~He, ``Non-local neural networks,'' in
  \emph{Proceedings of the IEEE conference on computer vision and pattern
  recognition}, 2018, pp. 7794--7803.

\bibitem{wang2020non}
Z.~Wang, N.~Zou, D.~Shen, and S.~Ji, ``Non-local {U}-{N}ets for biomedical
  image segmentation,'' in \emph{Thirty-Fourth AAAI Conference on Artificial
  Intelligence}, 2020.

\bibitem{vaswani2017attention}
A.~Vaswani, N.~Shazeer, N.~Parmar, J.~Uszkoreit, L.~Jones, A.~N. Gomez,
  {\L}.~Kaiser, and I.~Polosukhin, ``Attention is all you need,'' in
  \emph{Advances in Neural Information Processing Systems}, 2017, pp.
  5998--6008.

\bibitem{pei2020geom}
H.~Pei, B.~Wei, K.~C.-C. Chang, Y.~Lei, and B.~Yang, ``Geom-{GCN}: Geometric
  graph convolutional networks,'' in \emph{International Conference on Learning
  Representations}, 2020.

\bibitem{tenenbaum2000global}
J.~B. Tenenbaum, V.~De~Silva, and J.~C. Langford, ``A global geometric
  framework for nonlinear dimensionality reduction,'' \emph{science}, vol. 290,
  no. 5500, pp. 2319--2323, 2000.

\bibitem{nickel2017poincare}
M.~Nickel and D.~Kiela, ``Poincar{\'e} embeddings for learning hierarchical
  representations,'' in \emph{Advances in Neural Information Processing
  Systems}, 2017, pp. 6338--6347.

\bibitem{ribeiro2017struc2vec}
L.~F. Ribeiro, P.~H. Saverese, and D.~R. Figueiredo, ``struc2vec: Learning node
  representations from structural identity,'' in \emph{Proceedings of the 23rd
  ACM SIGKDD International Conference on Knowledge Discovery \& Data Mining},
  2017, pp. 385--394.

\bibitem{chollet2017xception}
F.~Chollet, ``Xception: Deep learning with depthwise separable convolutions,''
  in \emph{Proceedings of the IEEE conference on computer vision and pattern
  recognition}, 2017, pp. 1251--1258.

\bibitem{xu2018representation}
K.~Xu, C.~Li, Y.~Tian, T.~Sonobe, K.-i. Kawarabayashi, and S.~Jegelka,
  ``Representation learning on graphs with jumping knowledge networks,'' in
  \emph{International Conference on Machine Learning}, 2018, pp. 5449--5458.

\bibitem{li2018deeper}
Q.~Li, Z.~Han, and X.-M. Wu, ``Deeper insights into graph convolutional
  networks for semi-supervised learning,'' in \emph{Thirty-Second AAAI
  Conference on Artificial Intelligence}, 2018.

\bibitem{chen2019measuring}
D.~Chen, Y.~Lin, W.~Li, P.~Li, J.~Zhou, and X.~Sun, ``Measuring and relieving
  the over-smoothing problem for graph neural networks from the topological
  view,'' in \emph{Thirty-Fourth AAAI Conference on Artificial Intelligence},
  2020.

\bibitem{alon2020bottleneck}
U.~Alon and E.~Yahav, ``On the bottleneck of graph neural networks and its
  practical implications,'' \emph{arXiv preprint arXiv:2006.05205}, 2020.

\bibitem{sen2008collective}
P.~Sen, G.~Namata, M.~Bilgic, L.~Getoor, B.~Galligher, and T.~Eliassi-Rad,
  ``Collective classification in network data,'' \emph{AI magazine}, vol.~29,
  no.~3, pp. 93--93, 2008.

\bibitem{tang2009social}
J.~Tang, J.~Sun, C.~Wang, and Z.~Yang, ``Social influence analysis in
  large-scale networks,'' in \emph{Proceedings of the 15th ACM SIGKDD
  international conference on Knowledge discovery and data mining}, 2009, pp.
  807--816.

\bibitem{rozemberczki2019multi}
B.~Rozemberczki, C.~Allen, and R.~Sarkar, ``Multi-scale attributed node
  embedding,'' \emph{arXiv preprint arXiv:1909.13021}, 2019.

\bibitem{newman2002assortative}
M.~E. Newman, ``Assortative mixing in networks,'' \emph{Physical review
  letters}, vol.~89, no.~20, p. 208701, 2002.

\bibitem{gao2019graphattn}
H.~Gao and S.~Ji, ``Graph representation learning via hard and channel-wise
  attention networks,'' in \emph{Proceedings of the 25th ACM SIGKDD
  International Conference on Knowledge Discovery \& Data Mining}, 2019, pp.
  741--749.

\bibitem{knyazev2019understanding}
B.~Knyazev, G.~W. Taylor, and M.~Amer, ``Understanding attention and
  generalization in graph neural networks,'' in \emph{Advances in Neural
  Information Processing Systems}, 2019, pp. 4204--4214.

\bibitem{yang2016hierarchical}
Z.~Yang, D.~Yang, C.~Dyer, X.~He, A.~Smola, and E.~Hovy, ``Hierarchical
  attention networks for document classification,'' in \emph{Proceedings of the
  2016 Conference of the North American chapter of the Association for
  Computational Linguistics: Human Language Technologies}, 2016, pp.
  1480--1489.

\bibitem{krizhevsky2012imagenet}
A.~Krizhevsky, I.~Sutskever, and G.~E. Hinton, ``Imagenet classification with
  deep convolutional neural networks,'' \emph{Advances in neural information
  processing systems}, vol.~25, pp. 1097--1105, 2012.

\bibitem{lim2021new}
D.~Lim, X.~Li, F.~Hohne, and S.-N. Lim, ``New benchmarks for learning on
  non-homophilous graphs,'' \emph{arXiv preprint arXiv:2104.01404}, 2021.

\bibitem{hu2020open}
W.~Hu, M.~Fey, M.~Zitnik, Y.~Dong, H.~Ren, B.~Liu, M.~Catasta, and J.~Leskovec,
  ``Open graph benchmark: Datasets for machine learning on graphs,''
  \emph{Neural Information Processing Systems (NeurIPS)}, 2020.

\bibitem{leskovec2014snap}
J.~Leskovec and A.~Krevl, ``Snap datasets: Stanford large network dataset
  collection,'' 2014.

\bibitem{zhu2020beyond}
J.~Zhu, Y.~Yan, L.~Zhao, M.~Heimann, L.~Akoglu, and D.~Koutra, ``Beyond
  homophily in graph neural networks: Current limitations and effective
  designs,'' \emph{Advances in Neural Information Processing Systems}, vol.~33,
  2020.

\bibitem{jin2021node}
W.~Jin, T.~Derr, Y.~Wang, Y.~Ma, Z.~Liu, and J.~Tang, ``Node similarity
  preserving graph convolutional networks,'' in \emph{Proceedings of the 14th
  ACM International Conference on Web Search and Data Mining}, 2021, pp.
  148--156.

\bibitem{bo2021beyond}
D.~Bo, X.~Wang, C.~Shi, and H.~Shen, ``Beyond low-frequency information in
  graph convolutional networks,'' in \emph{Proceedings of the AAAI Conference
  on Artificial Intelligence}, vol.~35, no.~5, 2021, pp. 3950--3957.

\bibitem{zhu2021graph}
J.~Zhu, R.~A. Rossi, A.~Rao, T.~Mai, N.~Lipka, N.~K. Ahmed, and D.~Koutra,
  ``Graph neural networks with heterophily,'' in \emph{Proceedings of the AAAI
  Conference on Artificial Intelligence}, vol.~35, no.~12, 2021, pp.
  11\,168--11\,176.

\bibitem{paszke2017automatic}
A.~Paszke, S.~Gross, S.~Chintala, G.~Chanan, E.~Yang, Z.~DeVito, Z.~Lin,
  A.~Desmaison, L.~Antiga, and A.~Lerer, ``Automatic differentiation in
  pytorch,'' 2017.

\bibitem{Fey2019pyg}
M.~Fey and J.~E. Lenssen, ``Fast graph representation learning with {PyTorch
  Geometric},'' in \emph{ICLR Workshop on Representation Learning on Graphs and
  Manifolds}, 2019.

\end{thebibliography}
%



%




\end{document}